\documentclass[11pt]{article}

\usepackage[preprint]{acl}

\usepackage{times}
\usepackage{latexsym}

\usepackage[T1]{fontenc}

\usepackage[utf8]{inputenc}

\usepackage{microtype}

\usepackage{inconsolata}

\usepackage{graphicx}
\usepackage{booktabs}

\usepackage{amsmath}

\definecolor{dataedit}{HTML}{007A7A}

\title{GAAC: Lossless KV Compression for Bandwidth-Efficient LLM Inference on Personal Devices}

\author{Yipin Guo, Arun M George, Siddharth Joshi  \\
  University of Notre Dame \\
  \texttt{yguo23, ageorg23, sjoshi2 \space @nd.edu}
  }

\begin{document}
\maketitle

\begin{abstract}
Weight-only, lossy quantization substantially reduces the storage and memory requirements for serving large language models (LLMs), but the growing key--value (KV) cache remains a bottleneck for long-context inference on personal devices.
We present Group-Adaptive Additive Codec (GAAC), a lossless KV compression and serving system that exploits exponent redundancy to reduce storage and accelerate attention without introducing KV approximation.
GAAC encodes BF16 exponents using calibrated per-channel base-plus-offset codes over contiguous ranges. GAAC adapts code widths across small groups while preserving KV values bit-for-bit.
A fused attention kernel reconstructs KV tiles on the fly, avoiding full-cache decompression.
Block-wise compression and fixed-capacity slots support incremental cache updates and paged serving.
Across five tested models, GAAC achieves 1.405--1.407$\times$ lossless BF16 KV compression.
We also study compression performance on low-precision tensors FP8 E4M3 caches ( 1.09--1.14$\times$) and E5M2 caches (1.25--1.30$\times$).
On NVIDIA DGX Spark, GAAC accelerates decode attention at a 64K-token context by 1.19--1.27$\times$ and end-to-end W4A16 decode steps by 1.04--1.15$\times$.
In serving, it reduces resident KV memory by approximately 26\% and increases aggregate throughput by 13--29\% under comparable KV-memory budgets.
On Apple M4, GAAC achieves 1.02--1.25$\times$ attention speedups, demonstrating portability beyond NVIDIA GPUs.
Code is available at \url{https://github.com/Intelligent-Microsystems-Lab/GAAC}.
\end{abstract}

\begin{figure*}[t]
    \centering
    \includegraphics[width=\textwidth]{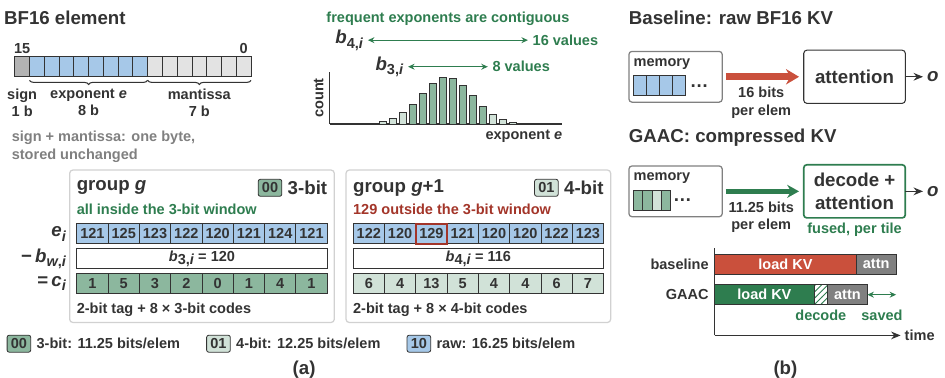}
    \caption{
    Overview of GAAC. 
    (a) GAAC exploits the concentration of BF16 exponent values in contiguous high-frequency regions. Each group of eight exponents shares a 2-bit tag indicating 3-bit, 4-bit, or raw mode. 
    The sign and mantissa bits of each element are stored unchanged in a single byte. Each code is an offset from a calibrated per-channel base, $c_i=e_i-b_{w,i}$. In the illustrated example, one exponent outside the 3-bit window causes the entire group $g{+}1$ to use 4-bit mode. Bits per element include the sign-mantissa byte and the mode tag.
    (b) Compared with standard attention, GAAC reduces KV loading time by reading fewer bytes from the compressed cache, with only a small computational overhead from on-the-fly decompression. $o$ denotes the attention output. Example values and timelines are illustrative.}
    \label{fig:GAAC_overview}
\end{figure*}

\section{Introduction}
\label{sec:intro}

Running large language models (LLMs) on personal devices requires fitting both model weights and the growing key-value (KV) cache within a limited memory budget. On personal platforms such as NVIDIA DGX Spark and Apple Silicon, decoding is typically constrained by memory bandwidth rather than available arithmetic throughput~\cite{flashdecoding++}.
Weight-only quantization~\citep{frantar2023gptq,guo2024gptqt,you2024shiftaddllm,guo2026qtea} has been the common, practical solution for this setting.
By reducing weight precision while retaining higher-precision activations, model storage and memory traffic from weight reads can be substantially reduced, with minimal quality.

However, shrinking the weights leaves the KV cache unchanged, increasing its relative contribution to memory consumption and decoding traffic.
KV storage and memory traffic become particularly costly in long-context applications: the cache grows with sequence length, and full attention repeatedly reads the cached keys and values during generation~\citep{hooper2024kvquant}.
Consequently, fitting a longer context into memory does not necessarily make processing that context fast.

KV pruning and quantization~\citep{zhang2023h2o,liu2024kivi} can reduce these costs, but introduce additional approximation and possibly performance loss~\cite{dfloat11}.
For deployments already using aggressively quantized weights, this creates another source of error that requires validation across tasks and context lengths.
Instead, alternative methods to reduce KV storage and bandwidth requirements with minimal impact on quality would be critical to efficient serving of LLMs. 

Lossless compression offers one such method, but storage savings alone do not imply faster attention.
Prior lossless compression methods exploit floating-point redundancy to reduce model weight storage~\citep{dfloat11} or accelerate KV transfer between server GPUs~\citep{guo2026splitzip}.
However, accelerating long-context attention, has different needs, compressed KV must remain resident in device memory and be reconstructed on demand, avoiding full-cache decompression and its associated memory traffic.
On bandwidth-constrained devices, the codec and attention kernel  must therefore be co-designed to be memory-layout aware.

We introduce \textbf{G}roup-\textbf{A}daptive \textbf{A}dditive \textbf{C}odec (\textbf{GAAC}), a lossless KV compression system for bandwidth-efficient inference on personal devices.
GAAC encodes BF16 exponents using short codes adapted to each channel, increasing the code width for groups that require more bits while preserving every sign, exponent, and mantissa bit.
The attention kernel reconstructs KV values on demand from compressed storage, trading lightweight computation for reduced memory traffic without materializing the full decompressed cache.
For serving, we compress the KV cache in blocks aligned with Tensor Core processing granularity. This supports efficient compression and memory management as new KV entries are added during decoding. Our contributions are:

\begin{itemize}
    \item \textbf{A bandwidth-oriented lossless KV codec.}
    We design a group-adaptive lossless BF16 KV compression method for devices with a high compute-to-memory-bandwidth ratio, balancing compact storage with efficient, bit-exact reconstruction.

    \item \textbf{Attention acceleration through reduced data movement.}
    We fuse KV reconstruction with attention, reducing memory traffic and reducing attention latency, especially during long-context decoding.

    \item \textbf{Efficient compressed KV cache management.}
    We design a block-wise compression strategy to minimize online encoding and decoding overhead. A slot-based memory management scheme accommodates variable-length compressed blocks within two fixed-size storage classes, enabling integration into existing LLM serving engines.
\end{itemize}

\section{Related Work}
\label{sec:related}

\subsection{Lossless LLM Weight Compression}

General-purpose GPU libraries such as nvCOMP~\citep{nvidia2026nvcomp} provide lossless compression primitives.
LLM-oriented methods further exploit the structure of floating-point model parameters.
DFloat11~\citep{dfloat11} uses variable-length exponent coding and GPU decompression to reduce BF16 weight storage without changing the original values.
ZipNN~\citep{zipnn} develops neural-network-specific lossless compression for reducing model storage and distribution costs.
Falcon~\citep{Li2025falcon} accelerates floating-point compression through GPU-oriented, precision-preserving conversion and adaptive bit-plane encoding.
ZipServ~\citep{fan2026zipserv} targets inference by combining Tensor-Core-Aware Triple Bitmap Encoding with a fused decompression-GEMM kernel that processes compressed weights without materializing an intermediate decompressed tensor.

These studies highlight the benefits of exploiting numerical redundancy and co-designing compression together with the kernels that use the compressed data.
Unlike static model weights, which can be compressed offline, KV states are generated incrementally during inference and must be compressed online as the cache grows. The compression scheme must therefore keep encoding overhead low, reduce cache storage and support efficient reconstruction within attention.

\subsection{KV Cache and State Compression}

Lossy KV compression primarily reduces either the number of cached tokens or the numerical precision of the stored KV.
H$_2$O~\citep{zhang2023h2o} retains recent tokens and attention-heavy hitters while evicting less important KV entries; SnapKV~\citep{li2024snapkv} uses attention patterns within an observation window to identify important positions for each attention head.
For quantization, KVQuant~\citep{hooper2024kvquant} combines per-channel, pre-RoPE key quantization with non-uniform representations and outlier handling to support low-bit KV storage.
These methods reduce KV memory by discarding tokens or approximating stored values, potentially introducing additional errors on top of weight quantization.

Lossless compression also reduces the volume of data transferred when communicating other model states.
ZipCCL~\citep{lin2026zipccl} compresses activations, gradients, and parameters exchanged by communication collectives during distributed training, targeting transfers between devices rather than the storage and memory traffic of the KV cache. 

SplitZip~\citep{guo2026splitzip} is a high-throughput codec targeting lossless KV compression.
It encodes frequent exponents with fixed 4-bit codes and handles rare values through an explicit escape stream, prioritizing encoding and decoding throughput for prefill–decode disaggregation.
However, it is designed to decompress the full cache once after transfer, and its format cannot be decoded piece by piece inside attention.

\section{Group-Adaptive Additive Codec}
\label{sec:GAAC}

GAAC couples a bandwidth-oriented lossless codec with compressed-cache attention and block-wise cache management.
It keeps historical KV states compressed in memory and reconstructs them only when needed by attention.

\subsection{Lossless KV Compression}
\label{sec:GAAC_codec}

\begin{figure}[h]
    \centering
    \includegraphics[width=\columnwidth]{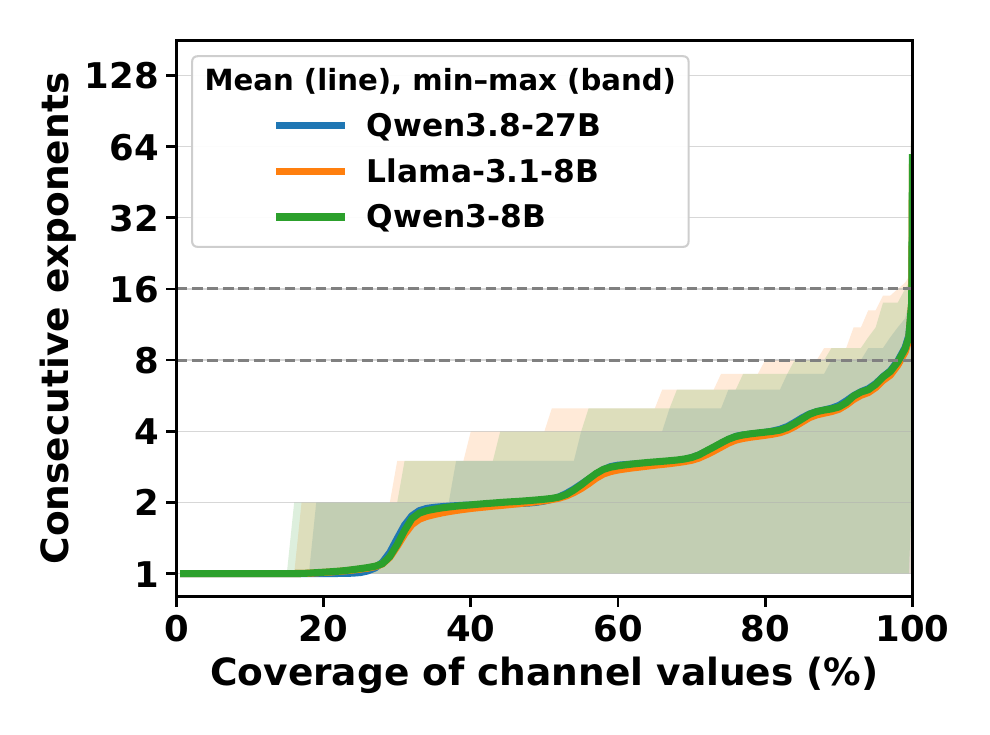}
    \vspace{-2.em}
    \caption{Concentration and contiguity of BF16 KV exponents. Most channels require fewer than eight consecutive exponents to cover 95\% of their values.}
    \label{fig:window_curve_combined}
\end{figure}

\paragraph{Contiguous exponent concentration.}
A BF16 element contains one sign bit, an 8-bit exponent, and a 7-bit mantissa. Consistent with prior observations on lossless compression~\citep{guo2026splitzip,dfloat11,fan2026zipserv}, BF16 exponents in LLMs typically have substantially lower entropy than their allocated bit width, whereas the mantissa fields contain little exploitable redundancy. We therefore store each element's sign and mantissa unchanged in a single byte and compress only its exponent.

Our KV analysis reveals that frequently occurring exponents are highly concentrated within short contiguous intervals, as illustrated in Figure~\ref{fig:window_curve_combined}. Across different LLM architectures, approximately 95\% of exponent values can be covered by only 6 consecutive exponents on average. Moreover, contiguous intervals achieve nearly the same coverage as selecting the most frequent exponents without contiguity constraints, motivating our base-plus-offset encoding scheme.

This structure allows us to replace an arbitrary exponent codebook with a base and a small integer offset. For each head–channel pair, calibration selects the contiguous windows of lengths 8 and 16 that cover the most exponent values, separately for K and V in each layer.
We denote their bases for element $i$ by $b_{3,i}$ and $b_{4,i}$.
These bases are shared across tokens in each KV cache block rather than fitted during every compression operation.
    
\paragraph{Group-adaptive additive coding.}
As shown in Figure~\ref{fig:GAAC_overview}(a), we partition KV elements into groups of eight in attention-access order.
Each group shares a 2-bit mode tag selecting 3-bit offsets, 4-bit offsets, or raw 8-bit exponents.
Writing $d_{w,i}=e_i-b_{w,i}$, the exponent width of group $g$ is
\begin{equation}
    w_g =
    \begin{cases}
        3, & 0 \le d_{3,i} < 8,\ \forall i \in g,\\
        4, & 0 \le d_{4,i} < 16,\ \forall i \in g,\\
        8, & \text{otherwise}.
    \end{cases}
    \label{eq:GAAC_group_mode}
\end{equation}

GAAC handles out-of-window exponents by increasing the code width for the entire group, avoiding individually indexed exceptions.
With $b_{8,i}=0$, encoding and reconstruction are
\begin{equation}
    c_i=e_i-b_{w_g,i},
    \qquad
    \widehat{e}_i=b_{w_g,i}+c_i.
    \label{eq:GAAC_additive}
\end{equation}
Decoding therefore employs integer addition rather than exponent-codebook lookups, making it better suited for personal devices with a high computation-to-bandwidth ratio.

All modes retain the low three bits of each code in a common packed stream.
A 4-bit group stores its eight additional high bits in one extra byte; a raw group stores the remaining five bits per exponent in five extra bytes.
The mode tag determines how many additional bytes to read and whether to add a calibrated base.
Grouping limits metadata to $2/8=0.25$ bits per element.
Therefore the common 3-bit mode occupies $8+3+0.25=11.25$ bits per element.
Because every exponent is reconstructed exactly and all sign and mantissa bits are retained, the original BF16 bit patterns are preserved.
Calibration mismatch affects compression efficiency, not correctness: values outside both windows use the raw mode.

\subsection{Attention with On-the-Fly Decompression}
\label{sec:GAAC_attention}

Compressing a cache but fully expanding it before attention would reintroduce full-size memory writes and reads.
GAAC instead fuses reconstruction with decode attention, as shown in Figure~\ref{fig:GAAC_overview}(b). The attention kernel reads compressed KV tiles and their metadata, reconstructs the required BF16 values, and immediately uses them to compute attention scores and weighted sums of values.
Only the current working tiles are reconstructed; no full-size decompressed cache is materialized in device memory.
Compressed blocks and any raw tail contribute to the same softmax normalization, preserving the original attention formulation.

This design trades lightweight reconstruction for reduced KV memory traffic.
The benefit is most pronounced in long-context decoding, where reading historical KV states can dominate attention latency and the saved memory-access time can outweigh reconstruction overhead.

\subsection{Block-Wise Compression Scheduling}
\label{sec:GAAC_scheduling}

\begin{figure}[h]
    \centering
    \includegraphics[width=\columnwidth]{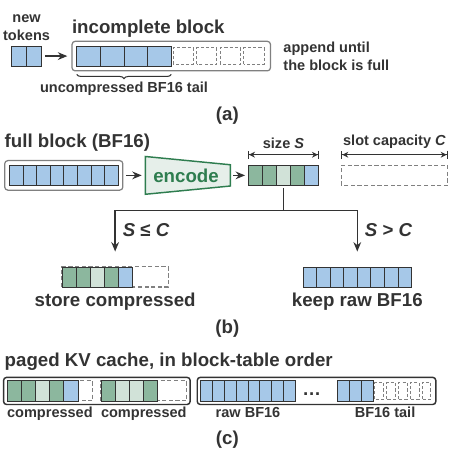}
    \caption{System design of GAAC.
    (a) New tokens are appended to an uncompressed BF16 tail until the block is full. (b) The full block is encoded once. If its compressed size $S$ does not exceed the slot capacity $C$, it is stored in a compressed slot; otherwise, it remains in BF16 format. (c) The paged KV cache holds compressed and raw finalized blocks in block-table order, followed by the incomplete BF16 tail.}
    \label{fig:GAAC_system}
\end{figure}

\begin{table*}[t]
    \centering
    \small
    \caption{Lossless KV compression across models. Entropy denotes the empirical entropy of the BF16 exponent stream. Ratio is the BF16 size divided by encoded size. Mode percentages are measured over 8-element groups.}
    \label{tab:compression_generality}
    \begin{tabular}{lcccccc}
        \toprule
        Model & $n_{kv}$ & KV exp. entropy (bit) & Ratio & 3-bit & 4-bit & Raw \\
        \midrule
        Qwen3-4B            & 8 & 2.78 & 1.405$\times$ & 89.0\% & 10.8\% & 0.2\% \\
        Qwen3-8B            & 8 & 2.78 & 1.405$\times$ & 88.9\% & 10.9\% & 0.2\% \\
        Qwen3.8-27B         & 4 & 2.69 & 1.405$\times$ & 88.3\% & 11.5\% & 0.2\% \\
        Qwen3.6-35B-A3B     & 2 & 2.71 & 1.405$\times$ & 88.4\% & 11.4\% & 0.2\% \\
        Llama-3.1-8B        & 8 & 2.70 & 1.407$\times$ & 89.9\% & 9.9\%  & 0.2\% \\
        \bottomrule
    \end{tabular}
\end{table*}

Inspired by paged KV management~\citep{kwon2023pagedattention}, we align compression units with token blocks, with K and V stored independently for each head.
Each block follows a \emph{fill-then-finalize} lifecycle.
New tokens are appended to an uncompressed BF16 tail while the block is incomplete.
When its last token arrives, the completed block is encoded once, and its compressed segments become the persistent cache representation.
After successful compression, the memory holding the uncompressed block is released for reuse, and finalized blocks are not recompressed as the sequence grows. During prefill, already complete blocks can be encoded directly, leaving only the incomplete suffix uncompressed.

The two levels of fallback in our codec produce variable-length compressed blocks, complicating memory allocation and management. We use fixed-capacity slots for compressed blocks and calibrate their capacity to maximize the compression ratio, balancing unused slot space against the frequency of raw-block fallbacks. If the encoded size of a block exceeds its slot capacity, the block remains in its original BF16 representation.
Consequently, the serving system only manages two physical storage classes: compressed slots and raw BF16 blocks.
The block table records the storage mode and physical location of each block, allowing attention to access both representations transparently.
This design preserves losslessness while supporting efficient memory management without arbitrary variable-size allocations. 
The storage analysis in Appendix \ref{sec:appendix_metadata} shows that fixed-capacity slots leave very little unused space. 

Deferring compression until block completion amortizes encoding over multiple token appends and avoids repeatedly repacking partially filled blocks.
Historical blocks remain compressed across subsequent decoding steps, reducing resident KV storage and repeated attention traffic.
The fused path targets decoding; prefill retains standard attention, temporarily expanding cached context when required.

\section{Experiments}
\label{sec:experiments}

\subsection{Setup}
\label{sec:exp_setup}

\paragraph{Hardware.}
We primarily evaluate GAAC on an NVIDIA DGX Spark equipped with a GB10 GPU, 48 SMs, and 128\,GB of unified LPDDR5x memory. The measured memory bandwidth is approximately 240\,GB/s.
We additionally evaluate the attention implementation on an Apple M4 with a 10-core GPU and 16\,GB of unified memory to study the portability of the proposed design across unified memory architectures.

\paragraph{Models and workloads.}
We evaluate dense models with 4B to 32B parameters from the Qwen3 and Llama-3.1 families~\citep{yang2025qwen3,qwen2025qwen25,grattafiori2024llama3}.
We also evaluate newer Qwen models with hybrid attention or mixture-of-experts (MoE) layers: Qwen3.8-27B and Qwen3.6-35B-A3B. Experiments use AWQ-quantized checkpoints~\citep{lin2024awq} with 4-bit weights and 16-bit activations (W4A16). We calibrate exponent windows on a subset of the WikiText~\citep{merity2017wikitext} training data. We test calibration transfer to code, math, knowledge, and language-modeling tasks with HumanEval~\citep{chen2021humaneval}, GSM8K~\citep{cobbe2021gsm8k}, MMLU~\citep{hendrycks2021mmlu}, and PTB~\citep{marcus1993ptb}.

\paragraph{Baselines and metrics.}
We compare GAAC with five GPU lossless codecs at the block level: nvCOMP~\citep{nvidia2026nvcomp}, DietGPU~\citep{johnson2022dietgpu}, DFloat11~\citep{dfloat11}, ZipServ~\citep{fan2026zipserv} and SplitZip~\citep{guo2026splitzip}.
For decode attention, we compare GAAC with FlashAttention~\citep{dao2022flashattention,dao2024flashattention2} using a BF16 KV cache.
For serving, we use the same vLLM~\citep{2026vllm} 0.17.1 engine with a BF16 paged KV cache as the baseline. Decode-step latency comparisons use the same model, batch size and context length. The compression ratio is the BF16 size divided by the complete compressed size, including mode tags and raw fallbacks.
Speedup is the baseline latency divided by the GAAC latency.
End-to-end serving accounts for online compression overhead and all other system overheads.

\subsection{Compression Effectiveness and Generality}
\label{sec:exp_generality}
The exponent structure exploited by GAAC remains consistent across model families, sizes, and attention architectures (Table~\ref{tab:compression_generality}). The empirical entropy of the BF16 exponent stream remains low across all five models, ranging from 2.69 to 2.78 bits.
As a result, 88.3--89.9\% of the 8-element groups use the 3-bit mode, 9.9--11.5\% use the 4-bit mode, and at most 0.3\% fall back to raw exponents.
GAAC reaches a compression ratio of 1.405--1.407$\times$ on every model. Accounting for the storage overhead of fixed-capacity slots, the effective reduction in resident KV memory is approximately 26\%, as detailed in Appendix \ref{sec:appendix_metadata}.
This approaches the 1.422$\times$ compression ratio of a cache encoded entirely in the 3-bit mode ($16/11.25$), showing that wider modes add little storage overhead. GAAC maintains consistent compression ratios across dense, hybrid, and MoE models with different numbers of KV heads.

We also evaluate how much additional lossless compression GAAC provides on FP8 KV caches. GAAC achieves 1.09--1.14$\times$ compression on E4M3 and 1.25--1.30$\times$ on E5M2. Detailed results are provided in Appendix \ref{sec:exp_fp8}.

\subsection{Block-Level Compression Latency}
\label{sec:exp_codec_latency}

\begin{figure}[h]
    \centering
    \includegraphics[width=0.8\columnwidth]{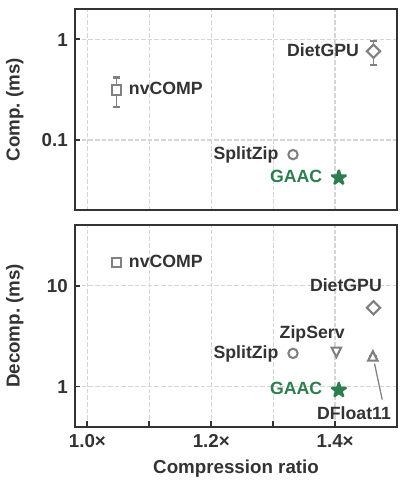}
    \vspace{-0.5em}
    \caption{Compression ratio and codec latency on KV states from one layer of Qwen3.8-27B. Compression latency is measured for a single KV block; decompression latency is measured for a 64K-token cache.}
    \label{fig:block_codec_baseline}
\end{figure}

We compare compression latency for a single KV block and decompression latency for a 64K-token cache, using KV states from one layer of Qwen3.8-27B. We evaluate DFloat11 and ZipServ only for decompression, as both are designed for fast decoding.

As shown in Figure~\ref{fig:block_codec_baseline}, GAAC achieves a compression ratio of approximately 1.405$\times$ with substantially lower encoding and decoding
latency than competing codecs. Compared with SplitZip and nvCOMP, it provides both
higher compression and faster processing. Although DietGPU achieves a slightly higher compression ratio, it incurs significantly greater latency.
These results show that additive coding combines compact storage with low decoding latency, enabling efficient on-the-fly KV reconstruction within attention.
\subsection{Long-Context Attention Acceleration}
\label{sec:exp_attention}

\begin{figure}[h]
    \centering
    \includegraphics[width=0.9\columnwidth]{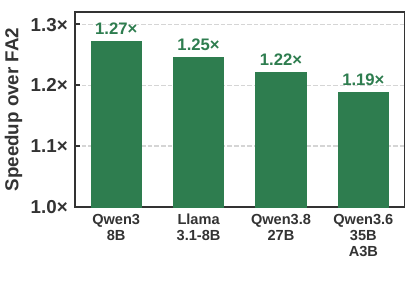}
    \vspace{-1.5em}
    \caption{Decode-attention speedup of GAAC using compressed KV over FlashAttention-2 using BF16 KV on DGX Spark, with a batch size of 1 and a 64K-token context.}
    \label{fig:attention_microbench}
\end{figure}

We next measure decode attention separately to determine whether the time saved by reducing KV traffic outweighs the overhead of on-the-fly reconstruction. Figure~\ref{fig:attention_microbench} compares GAAC on compressed KV with FlashAttention-2 on BF16 KV at a batch size of 1 and a 64K context.
GAAC is 1.19--1.27$\times$ faster on all four models.
The dense Qwen3-8B and Llama-3.1-8B gain the most, at 1.27$\times$ and 1.25$\times$.
GAAC achieves 1.19--1.22$\times$ speedups on the hybrid and MoE models despite their smaller per-layer KV data volume.

The speedup also holds across context lengths and batch sizes.
On Qwen3-8B, GAAC is 1.31$\times$, 1.27$\times$, and 1.25$\times$ faster at 32K, 64K, and 128K, and 1.29$\times$ faster with two 64K sequences. GAAC translates the 1.41$\times$ reduction in KV data volume into substantial attention speedups, with memory-access savings outweighing the overhead of on-the-fly reconstruction.


\subsection{End-to-End Serving}
\label{sec:exp_serving}

\begin{figure}[h]
    \centering
    \includegraphics[width=0.9\columnwidth]{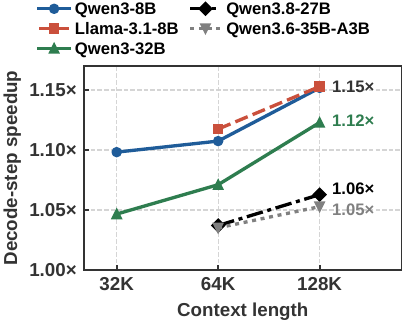}
    \vspace{-0.5em}
    \caption{End-to-end W4A16 decode-step speedup of GAAC over the BF16-KV baseline on DGX Spark at a batch size of 1.}
    \label{fig:e2e_decode}
\end{figure}

\begin{table*}[t]
    \centering
    \small
    \caption{Serving throughput under approximately equal KV-memory budgets.}
    \vspace{-0.5em}
    \label{tab:equal_memory}
    \begin{tabular}{lccrrrrc}
        \toprule
        & & & \multicolumn{2}{c}{BF16 KV} & \multicolumn{2}{c}{GAAC} & \\
        \cmidrule(lr){4-5} \cmidrule(lr){6-7}
        Model & Context & Budget & Batch & tok/s & Batch & tok/s & Gain \\
        \midrule
        Qwen3-8B    & 16K & 10 GB  & 4 & 60.3 & 5 & 68.1  & +13\% \\
        Qwen3-8B    & 64K & 20 GB & 2 & 19.0 & 3 & 22.8  & +20\% \\
        Qwen3-32B   & 64K & 35 GB & 2 & 8.47 & 3 & 10.51 & +24\% \\
        Llama-3.1-8B    & 64K & 18 GB & 2 & 20.0 & 3 & 24.9  & +25\% \\
        Qwen3.8-27B     & 64K & 14 GB & 3 & 23.3 & 4 & 29.6  & +27\% \\
        Qwen3.6-35B-A3B & 64K & 5 GB  & 3 & 62.9 & 4 & 81.3  & +29\% \\
        \bottomrule
    \end{tabular}
\end{table*}

\paragraph{Equal-batch decode.}
We measure complete W4A16 decode-step latency at a batch size of 1, including model computation and KV-cache access (Figure~\ref{fig:e2e_decode}).
GAAC speeds up every model at every context length by 1.04--1.15$\times$.
The speedup increases with context length because KV reads account for a larger share of decode-step latency.
Qwen3-8B improves from 1.10$\times$ at 32K to 1.15$\times$ at 128K, and Qwen3-32B from 1.05$\times$ to 1.12$\times$.
Llama-3.1-8B reaches 1.15$\times$ at 128K.
The hybrid Qwen3.8-27B and MoE Qwen3.6-35B-A3B achieve 1.04--1.06$\times$ speedups, as their KV caches are small relative to their model weights and only 25\% of their layers use full attention.
In all cases GAAC also cuts resident KV memory by about 26\%, for example from 19.3~GB to 14.2~GB for Qwen3-8B at 128K.

\paragraph{Equal-memory throughput.}
By reducing KV storage, GAAC also allows more sequences to be served concurrently within the same memory budget. We fix the KV-memory budget and increase the number of concurrent long-context sequences to the maximum each system can accommodate (Table~\ref{tab:equal_memory}).
Under the same budget, GAAC fits one more sequence for every model, and aggregate throughput rises by 13--29\%.
The largest gains appear on the hybrid Qwen3.8-27B (+27\%) and the MoE Qwen3.6-35B-A3B (+29\%), even though their equal-batch speedups are the smallest.
For these models, the main serving benefit comes from using the saved KV memory to support more concurrent sequences.

\subsection{Apple Silicon}
\label{sec:exp_apple}

\begin{figure}[h]
    \centering
    \includegraphics[width=0.9\columnwidth]{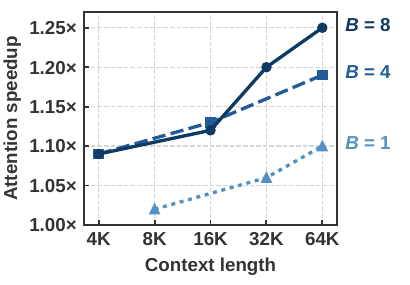}
    \vspace{-0.5em}
    \caption{Decode-attention speedup of GAAC’s Metal implementation over a BF16-KV kernel on Apple M4.}
    \label{fig:apple_m4}
\end{figure}

To test portability beyond NVIDIA GPUs, we implement the same compressed-KV attention path in Metal.
On an Apple M4, we compare it with a BF16-KV kernel of the same structure, using KV states generated by Qwen3-8B.
Figure~\ref{fig:apple_m4} shows that the GAAC attention kernel is 1.02--1.25$\times$ faster than the BF16-KV baseline across all ten configurations. The speedup grows with both context length and batch size.
With 8 sequences it rises from 1.09$\times$ at 4K to 1.25$\times$ at 64K, and with 4 sequences from 1.09$\times$ to 1.19$\times$.
For a single sequence, GAAC achieves speedups of 1.02$\times$ at 8K and 1.10$\times$ at 64K. At short contexts, speedups are smaller than on DGX Spark because reconstruction accounts for a larger share of execution time on the M4 GPU. Longer contexts and larger batches require each kernel to read more KV data, increasing the benefit of reduced memory traffic.


\subsection{Ablation Study}
\label{sec:ablation}

\begin{table*}[t]
    \centering
    \small
    \caption{Cross-task calibration generalization. Paired columns compare calibration on WikiText-2 ($A\!\rightarrow\!B$) with oracle calibration on the target task ($B\!\rightarrow\!B$). The lengths of the calibration prompts are consistent.} 
    \label{tab:calibration_dataset}
    \begin{tabular}{lcccccccc}
        \toprule
        & \multicolumn{2}{c}{3-bit groups(\%)}
        & \multicolumn{2}{c}{4-bit groups(\%)}
        & \multicolumn{2}{c}{Raw groups(\%)}
        & \multicolumn{2}{c}{Ratio($\times$)} \\
        \cmidrule(lr){2-3} \cmidrule(lr){4-5} \cmidrule(lr){6-7} \cmidrule(lr){8-9}
        Evaluation task $B$
        & $A\!\rightarrow\!B$ & $B\!\rightarrow\!B$
        & $A\!\rightarrow\!B$ & $B\!\rightarrow\!B$
        & $A\!\rightarrow\!B$ & $B\!\rightarrow\!B$
        & $A\!\rightarrow\!B$ & $B\!\rightarrow\!B$ \\
        \midrule
        WikiText-2 / language modeling & 88.3 & 88.3 & 11.5 & 11.5 & 0.20 & 0.20 & 1.405 & 1.405 \\
        HumanEval / code               & 87.1 & 88.8 & 12.6 & 11.0 & 0.35 & 0.16 & 1.402 & 1.406 \\
        GSM8K / math                   & 87.5 & 89.0 & 12.3 & 10.9 & 0.24 & 0.14 & 1.404 & 1.406 \\
        MMLU / knowledge               & 87.5 & 87.8 & 12.2 & 11.9 & 0.28 & 0.28 & 1.403 & 1.404 \\
        PTB / language modeling        & 88.3 & 88.9 & 11.5 & 11.0 & 0.23 & 0.14 & 1.405 & 1.406 \\
        \bottomrule
    \end{tabular}
\end{table*}

\subsubsection{Calibration Dataset Generalization}
\label{sec:ablation_calibration_dataset}

We test whether exponent windows calibrated on one dataset generalize to other tasks.
Let $A$ denote the WikiText-2 training data used for calibration.
For each target task $B$, we compare cross-task calibration $A\!\rightarrow\!B$ with oracle calibration $B\!\rightarrow\!B$, reporting the proportions of groups in 3-bit, 4-bit, and raw modes and the resulting compression ratio.

Table~\ref{tab:calibration_dataset} shows that windows calibrated once on WikiText-2 transfer well to code, math, knowledge, and language-modeling tasks.
Cross-task calibration reduces the proportion of 3-bit groups by at most 1.7 percentage points, while raw groups remain at or below 0.35\%.
The compression ratio drops by at most 0.004$\times$ relative to oracle calibration, from 1.406$\times$ to 1.402$\times$ on HumanEval.
A single offline calibration pass is sufficient for most workloads.

\subsubsection{Nested Precision Modes}
\label{sec:ablation_precision_modes}

We compare different precision-mode configurations to understand the tradeoff between compression efficiency and decoding overhead. We measure compression latency for a 16-token block and decoding latency for a 64K-token cache, excluding transmission latency.

GAAC uses a 2-bit mode tag to select among 3-bit, 4-bit, and raw encoding. The unused tag value could support a fourth mode with 2-bit codes. However, adding this mode improves the compression ratio by only 2.3\% while increasing decoding latency to 2.8 times that of the 3+4+raw design (Table~\ref{tab:precision_ablation}). Removing the 3-bit mode reduces decoding latency but lowers the compression ratio by 7.3\%. The 3+4+raw design retains most of the compression benefit of the four-mode scheme at substantially lower decoding latency.

\begin{table}[h]
    \centering
    \small
    \setlength{\tabcolsep}{3pt}
    \caption{Precision-mode ablation on Qwen3.8-27B.}
    \label{tab:precision_ablation}
    \begin{tabular}{lrrr}
        \toprule
        Scheme & Ratio & Compress ($\mu$s) & Decode ($\mu$s) \\
        \midrule
        2+3+4+raw      & 1.439$\times$ & 42.6 & 526.7 \\
        3+4+raw (ours) & 1.406$\times$ & 42.1 & 187.1 \\
        4+raw          & 1.304$\times$ & 36.7 & 158.9 \\
        \bottomrule
    \end{tabular}
\end{table}

\subsubsection{Attention Time Breakdown}
\label{sec:ablation_attn_breakdown}

To identify the source of GAAC's attention speedup, we separately measure the time spent on KV loading and computation during decode attention over a 64K-token context.

Table~\ref{tab:attn_breakdown} shows that GAAC reduces KV loading time by 294.3~$\mu$s while adding 49.2~$\mu$s of computation. Total attention latency therefore decreases from 1.36~ms to 1.11~ms, yielding a 1.22$\times$ speedup.
The time saved on KV loading outweighs the overhead of on-the-fly reconstruction.

Decompressing the cache before attention adds the cost of reading and decompressing the compressed KV data while retaining the full KV loading and computation costs of standard attention. Fusing reconstruction with attention avoids this separate decompression step and reduces KV loading time.

\begin{table}[h]
    \centering
    \small
    \caption{Decode-attention time breakdown over a 64K-token context. Times are in $\mu$s. Decomp.-first decompresses the cache before running attention.}
    \label{tab:attn_breakdown}
    \begin{tabular}{lrrr}
        \toprule
        Component & FA2 & Decomp.-first & GAAC \\
        \midrule
        KV loading     & 1117.8 & 1117.8 & 823.5 \\
        Exponent decoding & -    &  930.8   &  - \\
        Computation    & 241.5  & 241.5 & 290.7 \\
        \midrule
        Measured total & 1359.3 & 2290.1 & 1114.2 \\
        \bottomrule
    \end{tabular}
\end{table}

\section{Conclusion}
\label{sec:conclusion}

We presented Group-Adaptive Additive Codec, a lossless KV compression system that reduces both the memory footprint and the memory traffic of long-context LLM inference on bandwidth-constrained devices.
By exploiting the concentration of BF16 exponents in contiguous ranges, GAAC preserves the original KV bit patterns while enabling lightweight on-the-fly reconstruction within attention.
Combined with fill-then-finalize compression and paged management of compressed and raw blocks, this representation translates storage savings into lower decoding latency and higher serving throughput.
Across the evaluated models, GAAC reduces KV storage by approximately 26\%, accelerates long-context attention by up to 1.27$\times$, and increases equal-memory serving throughput by up to 29\%.
GAAC accelerates decode attention on both Apple M4 (1.02--1.25$\times$) and NVIDIA DGX Spark (1.19--1.27$\times$), demonstrating portability across architectures. 



\section*{Limitations}

The effectiveness of GAAC depends on redundancy in the exponent fields of the stored KV format.
GAAC provides less additional lossless compression on FP8 KV caches, particularly E4M3, than on BF16 KV caches.
Exponent windows are calibrated offline and transfer well across the evaluated tasks. However, RoPE-rotated keys can exhibit positional drift when calibration contexts are shorter than deployment contexts.
The codec remains lossless under such shifts, but wider encoding modes and raw fallbacks reduce compression efficiency.

The serving design also trades some space efficiency for predictable allocation.
Performance gains depend on the workload: on-the-fly reconstruction is most beneficial when KV memory reads dominate attention latency, whereas shorter contexts and models with fewer full-attention layers benefit less.
Our fused kernel accelerates decoding; prefill uses standard attention and does not receive the same direct acceleration.
Our evaluation covers selected LLMs on DGX Spark and Apple M4 and does not include power or energy measurements.
\section*{Ethical Considerations}

GAAC is an inference-efficiency method and does not alter model weights, training data, or the reconstructed BF16 KV values.
It therefore introduces no additional approximation at the KV representation level, but it does not address the safety, privacy, bias, or reliability risks of the underlying language model and its application.
Efficiency improvements may also increase overall deployment volume, so lower per-request resource use should not be interpreted as a demonstrated reduction in aggregate environmental impact.


\bibliography{reference_verified}

\newpage

\appendix
\twocolumn[{%
\begin{minipage}{\textwidth}

    \centering
    \small
    \captionof{table}{
    Qwen3.8-27B performance under different weight and activation precisions, with GAAC results for comparison.
    }
    \label{tab:activation_quality}
    \setlength{\tabcolsep}{3.5pt}
    \begin{tabular}{llccccccc}
        \toprule
        Benchmark & Metric
        & W4A16
        & W4AFP8
        & W4A4
        & W4-GAAC
        & W3A16
        & W3AFP8
        & W3-GAAC\\
        \midrule
        GSM8K
        & Acc. (\%) & 96.1 & 96.0  & 95.4 & 96.1 & 95.8 & 94.5 & 95.8 \\
        MATH-500
        & Acc. (\%) & 84.0 & 83.4 & 83.2 & 84.0 & 78.8 & 77.8 & 78.8  \\
        AIME 2024
        & Acc. (\%) & 80.0 & 76.7 & 76.7 & 80.0 & 56.7 & 50.8 & 56.7 \\
        HumanEval
        & pass@1 (\%) & 97.0 & 95.1 & 93.9 & 97.0 & 90.8 & 88.6 & 90.8\\
        MBPP
        & pass@1 (\%) & 78.8 & 78.6 & 77.4 & 78.8 & 74.8 & 73.6 & 74.8 \\
        LiveCodeBench
        & pass@1 (\%) & 37.0 & 36.4 & 34.9 & 37.0 & 37.8 & 35.8 & 37.8 \\
        \bottomrule
    \end{tabular}

\end{minipage}
\vspace{1em}
}]

\section{Model Quality under Activation Quantization}
\label{sec:exp_activation_quality}


Our target deployment uses aggressive weight quantization to reduce memory traffic from weight reads while retaining higher-precision activations whenever possible.
To measure the additional quality loss from activation quantization, we evaluate Qwen3.8-27B on mathematical reasoning and code-generation benchmarks using W4A16, W4AFP8, W4A4, W3A16, and W3AFP8. W4-GAAC and W3-GAAC denote the use of activations compressed via GAAC.
All variants use the same prompt template, decoding configuration, and evaluation harness.

Table~\ref{tab:activation_quality} shows that reducing activation precision introduces additional quality degradation, particularly on challenging reasoning and coding tasks.
Under 4-bit weights, W4AFP8 achieves 76.7\% on AIME 2024 and 95.1\% on HumanEval, compared with 80.0\% and 97.0\% for W4A16.
More aggressive W4A4 quantization further reduces HumanEval accuracy to 93.9\% and LiveCodeBench pass@1 from 37.0\% to 34.9\%.
The degradation becomes more pronounced with 3-bit weights: W3AFP8 drops AIME 2024 accuracy from 56.7\% to 50.8\% relative to W3A16.
In contrast, W4-GAAC and W3-GAAC preserve the benchmark results of their respective A16 baselines while losslessly reducing KV storage to approximately 11.8 bits per element.
These results motivate combining aggressive weight quantization with lossless KV compression to reduce memory traffic without introducing additional numerical approximation.
\section{Relation to FP8 KV}
\label{sec:exp_fp8}

\begin{table}[h]
    \centering
    \small
    \caption{Lossless compressibility after FP8 KV quantization.}
    \label{tab:fp8_residual}
    \begin{tabular}{lcc}
        \toprule
        KV Format & Additional Ratio & Total vs. BF16 \\
        \midrule
        BF16      & 1.405--1.406$\times$ & 1.405--1.406$\times$ \\
        FP8 E4M3  & 1.09--1.14$\times$   & 2.18--2.28$\times$ \\
        FP8 E5M2  & 1.25--1.30$\times$   & 2.50--2.60$\times$ \\
        \bottomrule
    \end{tabular}
\end{table}

FP8 KV quantization reduces the storage cost of each element, while GAAC can further compress the quantized cache without introducing additional numerical error.
We evaluate lossless exponent coding on model-generated KV states converted to FP8 (Table~\ref{tab:fp8_residual}).
GAAC achieves an additional 1.09--1.14$\times$ compression on E4M3 and 1.25--1.30$\times$ on E5M2, corresponding to total compression ratios of 2.18--2.28$\times$ and 2.50--2.60$\times$ relative to BF16, respectively.
These results demonstrate that lossless KV compression complements FP8 quantization, providing further memory savings without any approximation beyond the initial FP8 conversion.
\section{RoPE Positional Drift}
\label{sec:ablation_rope}

\begin{table}[h]
    \centering
    \small
    \caption{Overflow storage in layer 5 of Qwen3-8B over a 32K-token prompt. For each successive 4K-token segment, we report the maximum across heads of the mean overflow bytes per block. Parentheses indicate calibration prompt lengths.}
    \label{tab:rope_drift}
    \begin{tabular}{lrrr}
        \toprule
        Segment & K ($\le$8K) & K (32K) & V ($\le$8K) \\
        \midrule
        0--4K   & 38  & 40 & 39 \\
        4--8K   & 37  & 37 & 38 \\
        8--12K  & 52  & 37 & 39 \\
        12--16K & 117 & 37 & 41 \\
        16--20K & 177 & 38 & 38 \\
        20--24K & 183 & 37 & 40 \\
        24--28K & 185 & 37 & 40 \\
        28--32K & 190 & 39 & 40 \\
        \bottomrule
    \end{tabular}
\end{table}

We isolate positional distribution shift by applying windows calibrated on short prompts to a 32K prompt.
Table~\ref{tab:rope_drift} shows that K overflow grows with position when the windows are calibrated on prompts of at most 8K tokens.
Mean overflow per tile stays at 37--38 bytes up to 8K, then rises to 117 bytes at 12--16K and 190 bytes at 28--32K, about five times the short-context level.
RoPE applies position-dependent rotations to keys, causing their exponents to shift outside windows calibrated on short prompts.
Values are not rotated, and their overflow remains at 38--41 bytes throughout the prompt.
Calibrating on 32K prompts removes the drift: K overflow then stays at 37--40 bytes in every segment.
We therefore calibrate on prompts as long as the deployment context.
\section{Storage}
\label{sec:appendix_metadata}

\begin{table}[h]
    \centering
    \small
    \caption{Storage breakdown for a compressed 16-token tile covering 128 head dimensions, measured on Qwen3.8-27B.}
    \label{tab:meta_tile}
    \begin{tabular}{lrr}
        \toprule
        Field & Bytes & Bits / elem. \\
        \midrule
        Sign and mantissa         & 2048 & 8.000 \\
        3-bit exponent codes      & 768  & 3.000 \\
        Mode tags (2 bits / group)& 64   & 0.250 \\
        Escape bytes (mean)       & 32   & 0.125 \\
        Block-table entry         & 4    & 0.016 \\
        Unused escape slot        & 96   & 0.375 \\ 
        Fallback raw tiles        & 4.4  & 0.017 \\
        Exponent bases (mean)      & 0.06 & 0.000 \\
        \midrule
        Total                     & 3016 & 11.783 \\
        \bottomrule
    \end{tabular}
\end{table}

Table~\ref{tab:meta_tile} details the storage layout of a compressed 16-token tile, with a 256-dimensional head represented as two 128-dimensional half-heads.
Beyond the unchanged sign and mantissa bits, each tile stores 3-bit exponent codes, 2-bit mode tags, and escape bytes for 4-bit and raw groups.
Additional overhead comes from fixed-capacity escape slots, block-table entries, and occasional raw fallbacks.
Including these costs, the average storage requirement is 3,016 bytes per tile, corresponding to 11.783 bits per element.
\section{Attention Baseline Selection}
\label{sec:appendix_flashinfer}

We compare GAAC with FA2 and FlashInfer in the 64K decode-attention setting of Section~\ref{sec:exp_attention} (Table~\ref{tab:flashinfer}). All methods are measured in the same process. We report the mean and standard deviation of the median per-call latency across five runs.
FlashInfer achieves similar latency to FA2 on DGX Spark, and GAAC is at least 1.20$\times$ faster than either.
FA3 and FA4 are not supported on DGX Spark. Its GB10 GPU (compute capability 12.1) lacks the Hopper warpgroup MMA and TMA paths used by FA3, as well as the \texttt{tcgen05} tensor-core instructions and tensor memory required by FA4 on data-center Blackwell. We therefore use FA2 as the FlashAttention baseline on this device.

\begin{table}[h]
    \centering
    \small
    \caption{Decode-attention baseline comparison over a 64K-token context on one Qwen3.8-27B layer. Speedup is the baseline latency divided by GAAC latency.}
    \label{tab:flashinfer}
    \begin{tabular}{lrr}
        \toprule
        Method & Latency ($\mu$s) & Speedup \\
        \midrule
        FA2 (block 256)            & 1359.3 $\pm$0.8 & 1.22$\times$ \\
        FlashInfer (page 16)       & 1343.4$\pm$0.5 & 1.21$\times$ \\
        FlashInfer (page 256)      & 1339.8$\pm$0.8 & 1.20$\times$ \\
        FA3, FA4                   & \multicolumn{2}{c}{not supported on GB10} \\
        GAAC                 & 1114.2$\pm$2.1  & -- \\
        \bottomrule
    \end{tabular}
\end{table}

\end{document}